\documentclass[
]{ceurart}

\usepackage{listings}
\usepackage{dirtytalk}
\usepackage{amsthm} 
\usepackage[most]{tcolorbox}
\usepackage{hyperref}  

\newcommand{\showdoiurl}[2]{%
  \ifx&#1&%
    \href{#2}{#2}%
  \else
    \href{https://doi.org/#1}{https://doi.org/#1}%
  \fi
}
\begin{document}


\conference{The World Conference on eXplainable Artificial Intelligence}

\title{Interpretable Neural System Dynamics: Combining Deep Learning with System Dynamics Modeling to Support Critical Applications}


\author[1]{Riccardo D'Elia}[%
orcid=0009-0009-6172-4049,
email=riccardo.delia@idsia.ch
]
\address[1]{University of Applied Sciences and Arts of Southern Switzerland, Dalle Molle Institute for Artificial Intelligence, Lugano, Switzerland}

\begin{abstract}
The objective of this proposal is to bridge the gap between Deep Learning (DL) and System Dynamics (SD) by developing an \textbf{interpretable neural system dynamics} framework. While DL excels at learning complex models and making accurate predictions, it lacks interpretability and causal reliability. Traditional SD approaches, on the other hand, provide transparency and causal insights but are limited in scalability and require extensive domain knowledge. To overcome these limitations, this project introduces a Neural System Dynamics pipeline, integrating Concept-Based Interpretability, Mechanistic Interpretability, and Causal Machine Learning. This framework combines the predictive power of DL with the interpretability of traditional SD models, resulting in both causal reliability and scalability. The efficacy of the proposed pipeline will be validated through real-world applications of the EU-funded AutoMoTIF project, which is focused on autonomous multimodal transportation systems. The long-term goal is to collect actionable insights that support the integration of explainability and safety in autonomous systems.
\end{abstract}


\maketitle

\section{Context and Motivation}

The field of System Dynamics (SD) has long focused on modeling complex systems that underpin many application domains.
In transportation logistics, for example, dynamical systems are used to model supply chain operations, traffic congestion, fleet management, and urban mobility planning.

Traditional SD models rely on differential equations and expert-defined rules to represent the evolution of a system over time. These models provide interpretable causal pathways and have long been valued for their transparency and accountability. However, they are constrained by simplifying assumptions that often fail to capture the full complexity of real-world systems and suffer from limited scalability as the number of interacting variables increases.

\begin{figure}[h]
  \centering
  \includegraphics[width=1\linewidth]{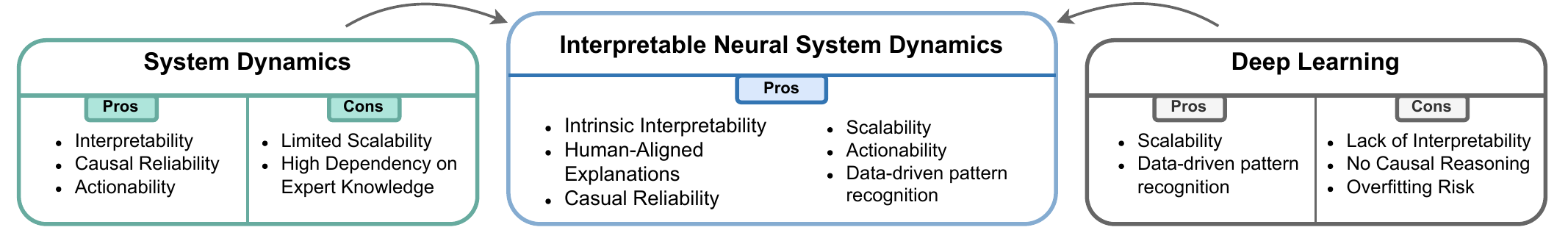}
  \caption{A unified framework combining System Dynamics and Deep Learning for Interpretable Neural System Dynamics.}
  \label{fig:INSD_framework}
\end{figure}

Contemporary Deep Learning (DL) techniques offer a promising alternative to overcome the aforementioned limitations of traditional SD modeling. DL algorithms are indeed capable of learning automatically the non-linear relationships that underpin dynamical systems' behaviors from large-scale data \cite{quaranta_review_2020}, thereby supporting the development of scalable and highly precise predictive models. 
Yet, these gains do not come costless.
Unlike traditional SD models, which involve concepts and inferential rules that are easily understandable by their users,
DL models operate as sort of \say{black boxes}, whose semantics and the decision-making logic behind their outputs remain mostly incomprehensible to users.
Furthermore, DL algorithms exploit predictions based on correlations, ignoring causal dependencies and mechanisms; this is referred to as a lack of \emph{causal reliability} \cite{illari_causality_2024}. Several methods have recently been proposed to overcome the opacity issues of DL models, which fall under the umbrella term of eXplainable AI (XAI). These methods provide valuable insight into how the DL models operate and produce their results. However, existing XAI techniques mostly fail to provide models with well-defined semantics that are interpretable to their users. This is because most available XAI methods are \emph{post-hoc}, inspecting the behaviors of naturally opaque models after training rather than trying to embed interpretable features directly within a model's structure.
Moreover, these methods are mostly incapable of addressing causal reliability problems as these fall beyond their usual target scope. This severely limits the usefulness of these methods in system dynamics, where interpretability and causal reliability are equally fundamental challenges that go hand in hand. Additionally, the difficulty of certifying deep learning systems due to their limited explainability poses significant safety concerns in critical applications \cite{flammini_towards_2024}.
This gap is increasingly being explored in the emerging field of \emph{Neuro-symbolic AI}, which integrates neural networks and symbolic reasoning to create interpretable and data-driven models. Recent advances in this field \cite{bhuyan_neuro-symbolic_2024} align with the goals of this research, which seeks to combine System Dynamics with Deep Learning in an attempt to get the best of both worlds, as detailed in Fig. \ref{fig:INSD_framework}.
This project offers a new approach to jointly address the challenges of interpretability and causal reliability posed by the widespread use of DL methods in the modeling of dynamical systems. Rather than focusing on the development and implementation of post-hoc interpretability techniques, the proposed approach focuses on the construction of a \textbf{interpretable by design neural systems dynamic} framework.
A plethora of different interpretability methods will be implemented to allow the combination of DL with the formalism of dynamical systems. 
In particular, the focus will be on techniques from the fields of \emph{Concept-Based Interpretability} \cite{poeta_concept-based_2023}, as well as \emph{Mechanistic Interpretability} \cite{bereska_mechanistic_2024} and \emph{Causal Machine Learning} \cite{kaddour_causal_2022}. The pipeline will be structured as follows, with a more detailed description provided in section \ref{section:pipeline}.

As a first step, concept-based interpretability methods will be employed to identify a set of semantically meaningful high-level variables (termed \say{concepts}) that describe understandable characteristics and magnitudes of interest. CML techniques will then be implemented to detect the causal dependencies among the selected high-level variables. Finally, mechanistically interpretable modeling techniques will be leveraged to infer a set of interpretable structural dynamic equations that govern the system's behavior. Such equations will be determined by taking into account the previously identified causal dependencies. This will not only allow for increased interpretability but will also contribute to anchoring the models to the real-world causal structure, making them substantially more reliable than existing DL algorithms. In turn, this enhanced reliability will also help address safety concerns, ensuring more robust and trustworthy AI systems.

As a final result, this pipeline should be able to return neural models that track the evolution of dynamical systems over time, both by operating on \textbf{semantically meaningful} and \textbf{actionable} variables. It will be implemented and evaluated on a real-world scenario from the EU-funded project \emph{AutoMoTIF}, where System Dynamics is involved in modeling the interoperability of multi-modal transportation terminals. A more detailed description of the real-world application is provided in Sec. \ref{section:automotif}.

\subsection{Real-World Application: EU Project AutoMoTIF}\label{section:automotif}

The doctoral research proposed herein will be carried out under the EU-funded \emph{AutoMoTIF} project\footnote{\url{https://automotif-project.eu}} (\emph{Automation towards multimodal transportation and integration of freight}). The project's core focus lies in the formulation of strategies, the development of business and governance models, and the generation of regulatory recommendations. These are designed to facilitate the integration and interoperability of automated transport systems. The project's overarching objective is to automate multimodal freight flows and logistics supply chains within the intra-European network, thereby enhancing operational efficiency and addressing existing regulatory and technological gaps.

Within this project, System Dynamics plays a crucial role in modeling and optimizing multimodal terminal operations, helping stakeholders to analyze and predict system behavior under different operational conditions. However, the complexity of these environments calls for data-driven AI approaches, which, despite their predictive power, often lack interpretability and causal reliability $-$ critical aspects for risk assessment and certification. Current deep learning models are opaque black boxes, making their decision processes difficult to understand and trust, particularly in safety-critical applications where accountability is essential. This challenge aligns with the broader Trustworthy AI paradigm, which underscores transparency, reliability, and human oversight \cite{bellogin_eu_2024}. Within this framework, the Trustworthy Autonomous Systems (TAS) research field focuses on developing methods to enhance AI accountability, explainability, and resilience in real-world deployments. Reflecting these concerns, the EU AI Act classifies AI-driven transport automation as a high-risk domain, requiring rigorous risk assessment, explainability, and robustness \cite{borrelli_applying_2024}. By integrating this research into \emph{AutoMoTIF}, the proposed pipeline will be tested in a real-world and high-pressure environment where understanding is crucial for ensuring safety, compliance, and trust in autonomous systems.

\section{The Many-Faces of the Interpretability Challenge}\label{section:interpretability_challenge}

Whilst the importance of eXplainable AI is becoming increasingly acknowledged, achieving interpretability remains a complex process.
The opacity of DL algorithms represents a major challenge for contemporary AI research.
In particular, the DL community must navigate various forms of opacity, which vary based on the different aspects of a model's structure and functioning that are focused on, as well as on the specific users and stakeholders involved \cite{sokol_explainability_2022}. 
The present project focuses specifically on two primary kinds of opacity that are of central relevance for research, notably \emph{semantic} and \emph{mechanistic} opacity \cite{muller_towards_2022}.

\textbf{Semantic opacity} refers to the challenge of deciphering what a model's learned representations mean in terms understandable to humans. This issue is particularly relevant in Neural Networks (NNs), where internal representations are often abstract and distributed without explicit meanings. Humans generally process information through high-level concepts, whereas NNs operate within a multi-dimensional feature space that obscures the semantics of the features involved and how these relate to categories comprehensible by the layman user. 

\textbf{Mechanistic opacity} refers to the difficulty of detailing precisely the mechanisms through which the various components of a model interact with one another and thus contribute to generating the overall model's behavior \cite{kastner_explaining_2024}. This issue is particularly significant in large-scale NNs with millions or even billions of parameters, where computations are spread across numerous layers and involve several non-linear transformations. 

In the XAI literature, these two different opacity forms have been mostly addressed separately by referring to different paradigms and implementing distinct strategies and techniques. This fragmented approach obstructs the creation of a cohesive, mathematically rigorous framework capable of addressing multiple interpretability challenges that stem from considering these two aspects of opacity together. Explaining the mechanisms underpinning the inference process of a DL model (e.g., via equation modeling) contributes minimally to the overall interpretability of the model's behavior if the features remain low-level and semantically meaningless. Conversely, mapping low-level features to high-level concepts has limited value if the model's decision-making mechanisms remain opaque.

\paragraph{Causal Reliability.} Related to the aforementioned forms of opacity is another fundamental issue, which is orthogonal to the problem of (mechanistic and semantic) opacity but intrinsically connected to it. This is the problem we refer to as \emph{causal reliability} \cite{illari_causality_2024}. This problem concerns the (in)ability of a model to track the real-world causal mechanisms operating beyond observable data generation and take them into account when drawing predictions. DL models are built to identify correlations among features and generate predictions solely based on them while ignoring the causal mechanisms. This poses DL algorithms in contrast with traditional mechanistic models, widely involved especially in the field of system dynamics. The latter, indeed, embeds an explicit representation of the causal mechanisms beyond data. The lack of reliance on real-world causal mechanisms represents a major limitation for DL algorithms, notably as it undermines their robustness and generalizability, especially in \emph{out-of-distribution} contexts \cite{scholkopf_toward_2021}. Furthermore, this issue limits the actionability of DL models, limiting the possibility of users intervening properly in their inferential processes and analyzing related interventional and counterfactual scenarios \cite{pearl_causality_2009}.

\paragraph{The Need for an Integrated Approach.} Opacity and causal reliability have been mostly treated as separate issues in contemporary AI research. Indeed, while opacity represents the target problem of XAI, causal reliability is at the heart of another growing research field, that of \emph{Causal Machine Learning} (CML) \cite{scholkopf_toward_2021, kaddour_causal_2022}. The two fields have developed separately with little connection among each other \cite{carloni_role_2023}. However, the two problems have a close relationship, especially when we focus on modeling dynamical systems. In response to these limitations, this doctoral project aims to propose a cohesive framework that jointly addresses the two aforementioned forms of opacity and, at the same time, produces causally reliable models. The project can be seen as an attempt to combine the three research domains of semantic (``concept-based'') explainability, mechanistic interpretability, and causal reliability, with specific reference to the field of dynamical systems modeling and its applications. These three domains will be combined into an integrated pipeline whose details are described in Sec. \ref{section:pipeline}.

\section{Research Strategy and Rationale}\label{section:pipeline}

\begin{figure}[h]
  \centering
  \includegraphics[width=1\linewidth]{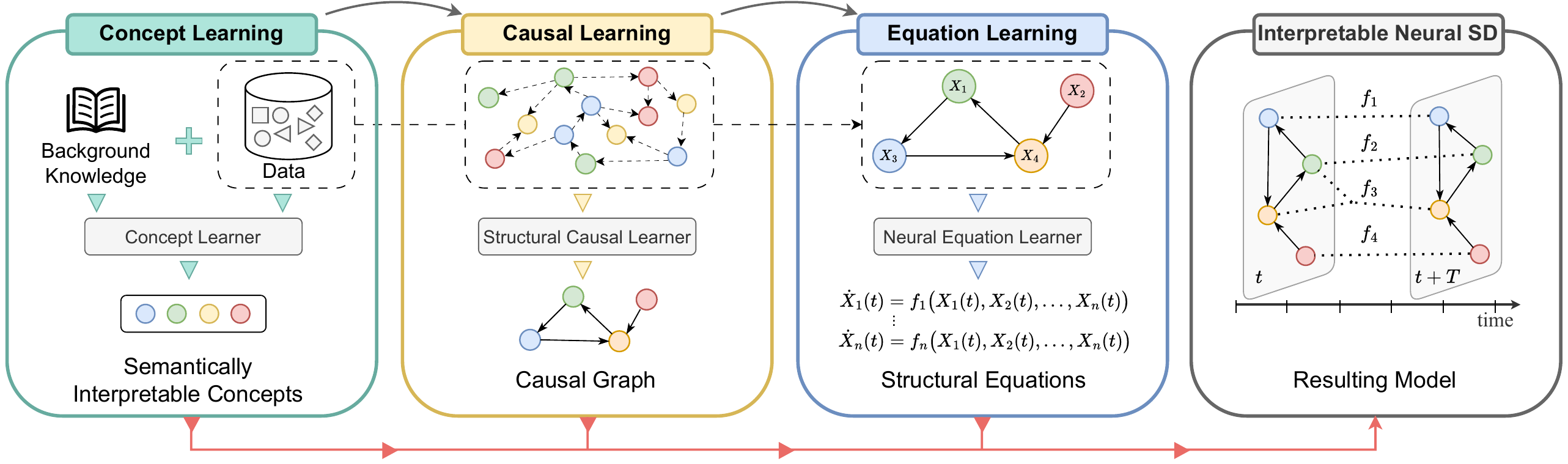}
    \caption{Overview of the INSD pipeline, from concept learning to causal and equation learning, ensuring interpretability and causal reliability in the resulting model.}
      \label{fig:INSD_pipeline}
\end{figure}

This project proposes a novel \emph{Interpretable Neural System Dynamics (INSD) pipeline} that combines Concept-Based Interpretability, Causal Learning, and Mechanistic Interpretability to construct causally-reliable neural system dynamics models that operate on human-interpretable variables while preserving the flexibility and scalability of DL approaches (Figure \ref{fig:INSD_pipeline}). The pipeline consists of three distinct learning steps:
\begin{enumerate}
    \item \emph{Concept Learning:} In the first step, concept-based interpretability (CBI) methods are used to extract high-level semantically interpretable variables (``concepts'') from raw data.
    \item \emph{Causal Learning:} In the second step, causal machine learning (CML) and causal discovery (CD) techniques are leveraged to identify the causal dependencies among these high-level concepts, thereby representing them in the form of a \emph{causal directed graph}.
    \item \emph{Equation Learning:} 
    In the third step, mechanistic interpretability methods are involved to derive explicit and interpretable dynamic equations that allow to predict the behavior of the target-system over time.
\end{enumerate}
After the three learning steps, the final model integrates the learned concepts, causal relationships, and governing equations to emulate the underlying dynamical system. Unlike traditional black-box neural networks, this model offers full interpretability, enabling users to trace predictions back to meaningful variables and causal influences. Furthermore, it provides actionable insights, allowing decision-makers to simulate interventions, predict long-term effects, and better understand the system’s behavior under different conditions. This ensures both transparency and practical applicability, bridging the gap between deep learning's flexibility and human-comprehensible system dynamics. The following paragraphs provide a detailed breakdown of each methodological component.

\subsection{Understanding System Dynamics Through Concept-Based Interpretability}
When applied to system dynamics, DL algorithms typically generate latent representations of system states that are usually not understandable and are arduous to interpret. For instance, in the context of epidemiological modeling, deep learning algorithms have the potential to discern underlying patterns of disease transmission; however, they have difficulty formulating these patterns using conventional epidemiological factors, such as \emph{contact rate} or \emph{incubation period}. This semantic opacity restricts their reliability for critical decision-making processes.
Concept-based Explainable AI introduces a potentially effective solution to these limitations by aligning AI reasoning with human-understandable abstractions rather than \emph{opaque} latent representations \cite{poeta_concept-based_2023}. Despite these innovations, this approach remains opaque regarding other aspects, such as the intrinsic mechanisms underlying concept representational learning. Concept-based models are currently underdeveloped concerning the temporal dimension, i.e. the evolution of interpretable concepts over time. This project would enable human interventions over evolving representations, constituting a significant advance.
In addition, this class of methods faces generalization and compositionality challenges because, similar to standard deep learning architectures, they are essentially associative models \cite{pearl_causality_2009}. This suggests that their decision-making process is not aligned with the underlying causal mechanisms of the world. They must distinguish regularities in data that reflect true causal relationships from those that are spurious. It is, therefore, crucial to comprehend this distinction to develop a robust and reliable understanding of phenomena, as well as to support intervention planning and ensure the application of fairness constraints \cite{scholkopf_toward_2021}.

\begin{tcolorbox}[colback=lightgray!30, colframe=black!50]
\paragraph*{Running Example: Automated Terminal Operation in \emph{AutoMoTIF}.}
Within an intermodal terminal setting, while a DL model might forecast freight congestion patterns, it may not clarify the reasons behind delays. By employing concept-based interpretability, logistics-related concepts such as terminal workload, handling efficiency, and waiting times for various transport modes are incorporated. This ensures predictions reflect real-world operational factors accurately. Consequently, this approach makes AI-driven simulations more transparent and actionable for terminal operators.
\end{tcolorbox}

\subsection{The Role of Causality in DL-based System Dynamics Models}

Causal reasoning plays a crucial role in System Dynamics, as these models explicitly represent causal connections between system variables. Contrarily, DL-based approaches rely on statistical correlations rather than true causal frameworks, which limits their ability to provide strong, interpretable predictions. A promising direction for overcoming these limitations is provided by recently developed CML techniques and, in particular, the framework of Neural Causal Models (NCMs). These techniques aim to uncover the underlying causal structure of a system by learning a graph that captures causal dependencies between concepts. Based on this learned graph, we can then infer equations that describe the system's evolution in a causally reliable manner, ensuring that the resulting models generalize more robustly and provide deeper insights into the underlying mechanisms governing the data.

\begin{tcolorbox}[colback=lightgray!30, colframe=black!50]
\paragraph*{Running Example: Automated Terminal Operation in \emph{AutoMoTIF}.}
A DL model might predict regular train delays at an intermodal terminal and identify a correlation between high truck traffic and these delayed departures. However, without causal reasoning, the underlying cause can remain unclear. A causal model could determine whether truck congestion is directly causing train delays or if another external factor, such as inefficient crane operation, is primarily responsible. By simulating counterfactual scenarios, such as \say{What if we increased crane availability?}, causal DL enables logistics operators to make proactive and data-driven decisions.
\end{tcolorbox}

\subsection{Understanding System Dynamics Through Mechanistic Interpretability}

Mechanistic Interpretability seeks to unravel DL models by reverse-engineering them to reveal their internal structures and decision-making processes. This research field is highly pertinent to System Dynamics, where comprehending a model's inner workings holds equal importance as its predictive accuracy. Traditional System Dynamics models use clearly defined equations and feedback loops, which make them inherently easy to interpret. The formal use of a causal loop diagram to describe a feedback system naturally leads to a connection with graph-based AI architectures like Graph Neural Networks (GNNs). GNNs offer an intuitive and organized way to represent dynamical systems, bringing benefits regarding intrinsic interpretability by exploiting relational inductive bias. Specifically, mechanistic interpretability techniques can help to trace information propagation to understand internal interactions. This is essential to develop structured, human-interpretable representations of dynamic processes.

\begin{tcolorbox}[colback=lightgray!30, colframe=black!50]
\paragraph*{Running Example: Automated Terminal Operation in \emph{AutoMoTIF}.}
Consider a simulation of an intermodal terminal where a DL-based model recommends rerouting trucks via a secondary access road. Without mechanistic interpretability, the reasoning behind this decision remains unclear $-$ whether it stems from predicted congestion, infrastructure constraints, or other operational factors. Taking advantage of the intrinsically interpretable model, we can identify modular components within the model, track the flow of information within the model, and uncover the key interactions influencing routing choices. This structured understanding enhances both the interpretability and trustworthiness of AI-driven logistics systems.
\end{tcolorbox}

\section{Research Questions and Objectives}

This research is guided by the following key questions:
\begin{itemize}
\item[\textbf{RQ1:}] How can a system dynamics framework ensure \emph{transparency} and \emph{accountability} while maintaining the predictive power of deep learning?

\textbf{Working Hypothesis}: Integrating traditional equation-based modeling with deep learning can achieve this balance.

\item[\textbf{RQ2:}] How can traditional modeling techniques and deep learning be effectively combined to leverage their strengths?

\textbf{Working Hypothesis}: An optimal integration may involve combining three key areas of XAI research: \emph{concept-based interpretability}, \emph{mechanistic interpretability}, and \emph{causal machine learning}.

\item[\textbf{RQ3:}] How can methods from the aforementioned distinct fields be integrated to develop an interpretable neural system dynamics framework?

\textbf{Working Hypothesis:} Concept-based interpretability can be used to learn high-level concepts and map raw data to meaningful variables, causal learning to infer dependencies among these variables, and mechanistic interpretability to define and parameterize system equations.
\end{itemize}

To refine this framework, the investigation focuses on: \emph{(i)} how concept-based techniques can identify high-level variables from raw data; \emph{(ii)} how causal learning can infer dependencies using data and background knowledge; and \emph{(iii)} how to determine the structure and parameters of governing equations in an interpretable manner. Additionally, the framework’s adaptability for modeling intermodal transportation logistics is examined, with an emphasis on ensuring safety and accountability. The integration of traditional equation-based modeling with deep learning, leveraging methods from these XAI fields, is hypothesized to achieve an interpretable and reliable system dynamics framework. This approach ensures \emph{partial verifiability, actionability, and control} in real-world applications.

\paragraph{Specific Objectives and Milestones.}

The project's \textbf{main objective} is to design, implement, and evaluate an integrated pipeline combining system dynamics modeling with deep learning, applying it to \emph{AutoMoTIF} and assessing generalisability.
To achieve this objective, the following intermediate milestones are planned:
\begin{itemize}
\item[\textbf{M1:}] State-of-the-Art Analysis. Review concept-based, mechanistic, and causal learning interpretability via a systematic literature review [\textbf{D1}: survey paper].

\item[\textbf{M2:}] Pipeline Blueprint. Design a blueprint for integrating dynamical system modeling with deep learning through literature review and theoretical modeling [\textbf{D2}: short conference paper].
\item[\textbf{M3:}] 

Use-Case Implementation. Develop and validate the pipeline on a toy example [\textbf{D3}: conference paper (e.g., NeurIPS, ICML)].

\item[\textbf{M4:}] 
Real-World Scalability Study. Scale the pipeline to \emph{AutoMoTIF} [\textbf{D4}: applied research paper].

\item[\textbf{M5:}] Generalization Study. Assess applicability to other real-world scenarios [\textbf{D5}: generalization study report or journal paper].
\item[\textbf{M6:}] Dissertation Completion. Compile research findings into the PhD thesis [\textbf{D6}: dissertation].
\end{itemize}

\begin{figure}[h]
  \centering
  \includegraphics[width=1\linewidth]{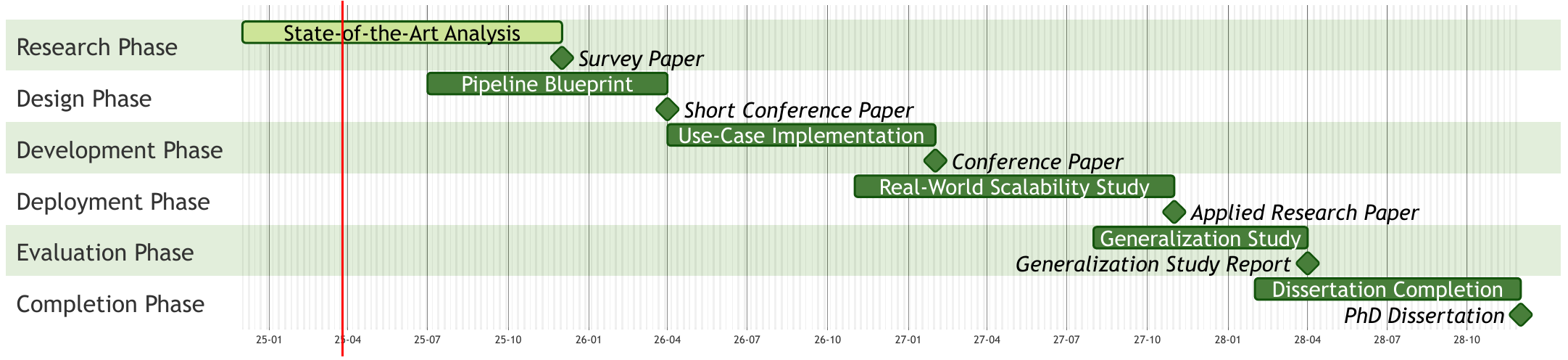}
    \caption{Gantt diagram of the doctoral project.}
      \label{fig:gantt}
\end{figure}

The project will run for 48 months. It started in January 2025, and termination is planned for December 2028. The structure of the project's timeline is reported in Fig. \ref{fig:gantt}.

\paragraph{Expected Contribution and Impact.}

Through the development of a unified interpretability framework for DL-based System Dynamics models, this research aspires to bridge the current divide between theoretical advancements in eXplainable AI and their application in high-stakes, real-world environments. By bringing together causal, mechanistic, and concept-based perspectives within a cohesive methodology, the project is expected to deliver not only novel algorithms and formal models but also practical tools that empower users to understand, trust, and effectively intervene in AI-driven processes.

The anticipated contributions extend beyond the specific context of multimodal logistics, offering generalizable insights for any domain that relies on the interplay of Deep Learning and System Dynamics. These insights may be applied to a wide range of fields, including, but not limited to, transportation, healthcare, environmental monitoring, and finance. Ultimately, this work aims to establish both a conceptual and operational foundation for interpretable, trustworthy AI in settings where transparency and accountability are not optional but essential for safety, compliance, and societal acceptance.

\begin{acknowledgments}
This work was partly supported by the Swiss State Secretariat for Education, Research and Innovation (SERI) under contract no. 24.00184 (AutoMoTIF project). The project has been selected within the EU Horizon Europe programme under grant agreement no. 101147693. Views and opinions expressed are however those of the authors only and do not necessarily reflect those of the funding agencies, which cannot be held responsible for them.
\end{acknowledgments}

\bibliography{DoctoralProposal_XAI_2025}


\end{document}